\documentclass{article}
\usepackage{spconf,amsmath,epsfig}
\usepackage{booktabs}
\usepackage{multirow}
\usepackage{adjustbox}
\begin{document}\sloppy

\def\x{{\mathbf x}}
\def\L{{\cal L}}

\title{FrameRank: A Text Processing Approach to Video Summarization}
%
\name{Zhuo Lei$^{1,2}$, Chao Zhang$^{1,2}$, Qian Zhang$^{2}$, Guoping Qiu$^{3,4}$}
\address{$^1$International Doctoral Innovation Center \\
$^2$School of Computer Science, The University of Nottingham Ningbo China \\
$^3$College of Information Engineering and Guangdong Key Lab for Intellignet Information Processing \\
Shenzhen University, China, $^4$School of Computer Science, The University of Nottingham, UK \\
\{Zhuo.LEI, Chao Zhang, Qian Zhang\}@nottingham.edu.cn, Guoping.Qiu@nottingham.ac.uk}

\maketitle

\begin{abstract}
Video summarization has been extensively studied in the past decades. However, user-generated video summarization is much less explored since there lack large-scale video datasets within which human-generated video summaries are unambiguously defined and annotated. Toward this end, we propose a user-generated video summarization dataset - UGSum52 - that consists of 52 videos (207 minutes). In constructing the dataset, because of the subjectivity of user-generated video summarization, we manually annotate 25 summaries for each video, which are in total 1300 summaries. To the best of our knowledge, it is currently the largest dataset for user-generated video summarization.

Based on this dataset, we present FrameRank, an unsupervised video summarization method that employs a frame-to-frame level affinity graph to identify coherent and informative frames to summarize a video. We use the Kullback-Leibler(KL)-divergence-based graph to rank temporal segments according to the amount of semantic information contained in their frames. We illustrate the effectiveness of our method by applying it to three datasets SumMe, TVSum and UGSum52 and show it achieves state-of-the-art results.
\end{abstract}
\section{Introduction}
\label{sec:intro}

User-generated video is growing exponentially. Hence, the demand for efficient ways of searching and retrieving desired content will cost huge amounts of resources like time, human resources and machine configurations. However, users always think little of time spending, cutting, content and view selection. Thus, user-generated videos consist of long, poorly-filmed (including illumination, shakiness, dynamic background and so on) and unedited contents. In this context, video summarization plays an important role in assisting users to quickly browse through important events contained in it. Recently, video summarization techniques have drawn a lot of attention especially for user-generated videos. The essential of user-generated video summarization is to identify important parts of original videos and define their importance. However, the insufficiency of publicly available datasets has limited this important line of research. Due to the subjectivity, human-generated summaries are the most needed to meet the purpose of training and evaluation for different user-generated video summarization methods. To help alleviate this, we introduce a new dataset, UGSum52, which contains 52 videos covering various user-generated contents and 25 human-generated video summaries for each one. Furthermore, another challenge is no standard criteria for measuring importance, even humans cannot agree on a universal basis for generating video summary. In this paper, we consider frame importance as the ones can most substitute others.

\begin{figure}[t]
\centering
\includegraphics[width=9cm]{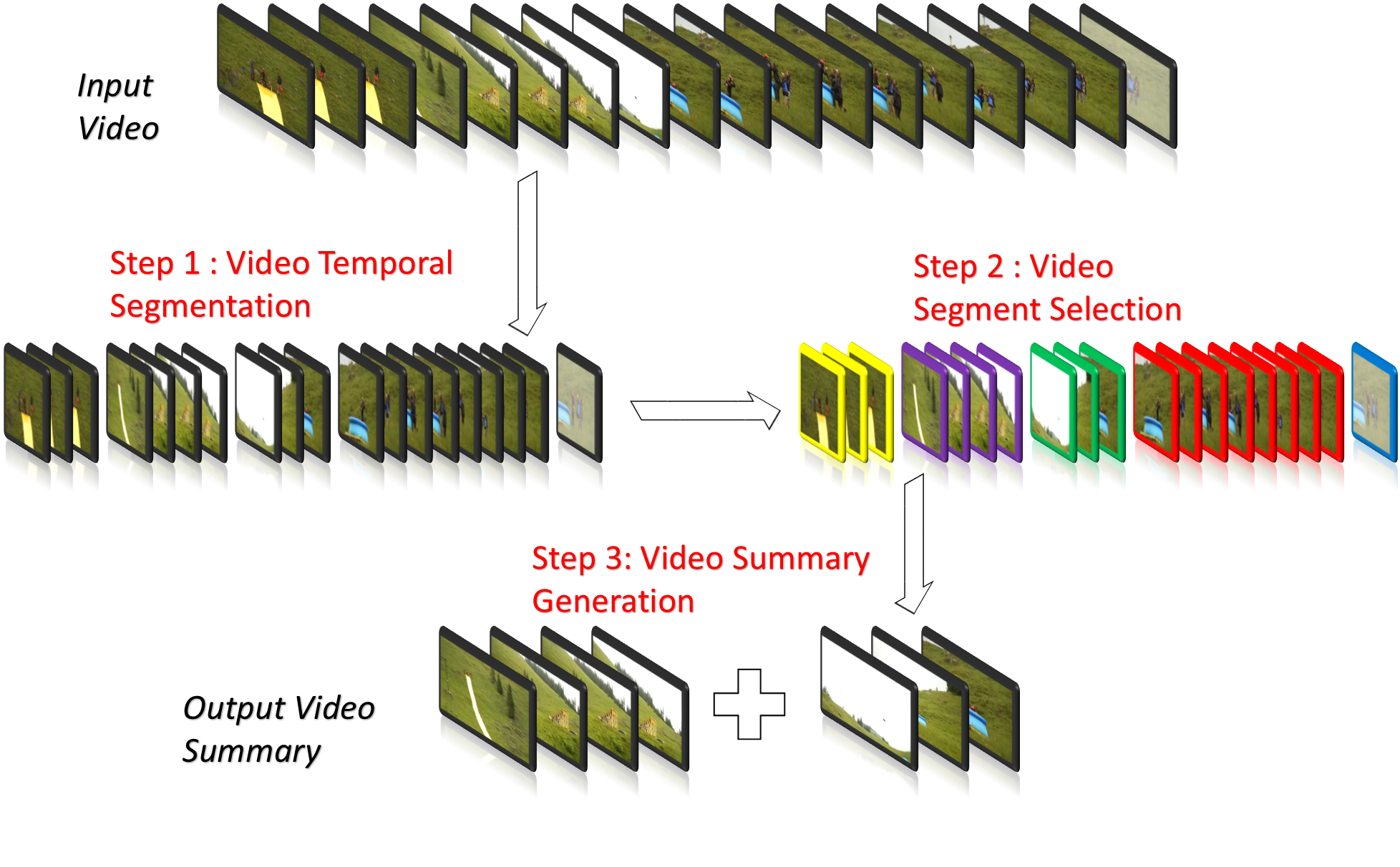}
\caption{Overview of the proposed framework.}
\label{Overview}
\end{figure}

Based on UGSum52 dataset, we propose a novel unsupervised framework for user-generated video summarization, which identifies coherent and informative video frames to summarize a video. As shown in Figure \ref{Overview}, we first divide an original video into disjoint segments with a dense-neighbor-based clustering method. We then develop a graph-based ranking method, FrameRank, to score and rank these segments according to the amount of semantic information. Finally, we sample video segments with high scores to generate video summaries. Through systematic experiments and evaluation, we show the proposed novel video summarization method is effective and outperforms state-of-the-art methods on the new UGSum52 dataset, and the other two existing datasets - SumMe \cite{Gygli} and TVSum \cite{Song}. Our main contributions are as the followings: 1) We introduce a new dataset for user-generated video summarization. To the best of our knowledge, it is the largest user-generated video summarization dataset able to meet the purpose of training and evaluation for different methods; 2) We develop a new method, FrameRank, to assess the importance of video frames; 3) We proposed a novel approach for video temporal segmentation, in which segments are semantically consistent and appropriate to produce good video summaries.

\section{Related Work}
\label{related}

\subsection{Datasets}

To facilitate the comparison between these datasets and our UGSum52 dataset, we show in Table \ref{5-dataset_related} more dataset statics. To be noted, the annotation information on UTE is not available since it does not apply human-generated video summaries to conduct the evaluation, while asking question to participants. Generally speaking, previous datasets have greatly boosted the researches in user-generated video summarization but still have several drawbacks. First, video types are still insufficient. Second, multiple human-generated video summaries for each video are necessary due to their subjectivity. Third, these datasets do not cover abundant categories.

\begin{table}
\scriptsize
\centering
\caption{Comparison between existing datasets.}
\begin{tabular}{|p{5.065em}|c|c|c|c|}
\toprule
\multicolumn{1}{|c|}{\textbf{Dataset}} & \textbf{UTE} & \textbf{SumMe} & \textbf{TVSum} & \multicolumn{1}{p{4.04em}|}{\textbf{UGSum52 (ours)}} \\
\midrule
Static \#Video & 0 & 4 & 41 & 2 \\
\midrule
Dynamic \#Video & 0 & 17 & 9 & 37 \\
\midrule
Egocentric \#Video & 4 & 4 & 0 & 13 \\
\midrule
Total \#Frame & 915980 & 109870 & 352356 & 812448 \\
\midrule
Avg. \#Frame & 228990 & 4395 & 7047 & 15624 \\
\midrule
Total Video Length (s) & 61065 & 4000 & 12582 & 12392 \\
\midrule
Avg. Video Length (s) & 15266 & 160 & 252 & 238 \\
\midrule
Avg. \#Summary per Video & N/A & 16 & 20 & 25 \\
\midrule
F-score & N/A & 0.31 & 0.36 & 0.31 \\
\midrule
Avg. Cronb. $\alpha$ & N/A & 0.71 & 0.81 & 0.78 \\
\bottomrule
\end{tabular}%
\label{5-dataset_related}%
\end{table}%

\subsection{User-generated Video Summarization}
Most works attempt to assess importance or interestingness of video frames with supervised methods. \cite{Lee} trained a model according to significant interaction between people and objects to learn the saliency in egocentric videos. \cite{Lu} built links between objects to create story-driven summaries grounded on \cite{Lee}'s egocentric feature. \cite{Gygli, GygliCVPR15} combined multiple features to train a regressor to predict interestingness. \cite{TingYao} devised a two-steam deep Convolutional Neural Network (CNN) architecture by fusing spatial and temporal information respectively on each steam for video highlight detection. \cite{ZhangKe} learned non-parametrically to transfer summary structures from training videos to the test ones, and \cite{KeZhangECCV2016} improved Long Short-Term Memory (LSTM) to model the variable-range temporal dependency among video frames. In general, supervised methods require a large amount of training data, which is what is lacking in this field. Existing datasets are not scalable to cover variety of user-generated videos, because what users are interested in cannot be exactly defined. Meanwhile, human-generated summaries are labor-intensive and time-consuming, and it requires multiple summaries for each single video due to the subjectivity. Therefore, the learned models may be not portable to work on user-generated videos.

Some unsupervised methods are put forward in user-generated video summarization. In detail, these methods use various types of intuitive criteria or pre-trained models from other fields to facilitate the assessment of importance or interestingness of video frames. With video's title or keywords as query, \cite{Khosla, Kim} obtained canonical viewpoints collected from website to predict important frames to achieve summarization. \cite{Song} employed an auto-encoder to train with internet videos of the same topic, and then assessed importance according to how well it can reconstruct input video's feature. \cite{Potapov} utilized a linear SVM classifier to obtain the confidence of event type as importance scores. However, although important frames can be effectively predicted, these methods can only work on domain-specific videos or require metadata of videos. Moreover, retrieving these images or videos is expensive, even collected metadata may not be relevant or correct.

\section{UGSum52 Benchmark Dataset}
\label{5-benchmark}

User-generated video summarization is a relatively unexplored domain, and there are few public datasets with multiple human-generated summaries available. We therefore collected a new dataset, UGSum52, that contains 52 videos and each has 25 human-annotated summaries.

\subsection{Video Collection}

\begin{figure}
\centering
\includegraphics[width=8cm]{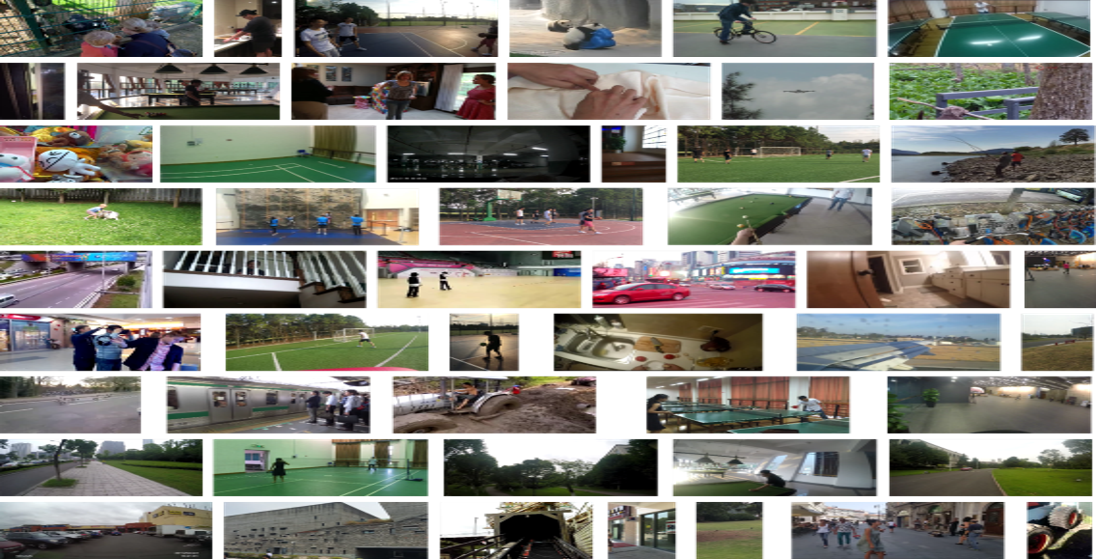}
\caption{We show these videos represented by their thumbnails.}
\label{datasetexam}
\end{figure}

We collected 52 videos captured either by ourselves or from YouTube, which are recorded in multiple ways, including static, dynamic and egocentric views. The duration ranges from 1 to 9 minutes. They are all raw or minimally edited. We collected videos encompassing various user-generated contents, like holidays, events and sports. Compared with other datasets, ours has more videos, categories and human-generated summaries. Figure \ref{datasetexam} shows the thumbnails.

\subsection{Video Annotation}
Due to the subjectivity of user-generated video summarization, it is almost impossible to obtain absolute ground truth labels, thus evaluation is often carried out with multiple human judgment. We asked 25 participants to collect 1,300 human generated summaries. Videos were shown to participants in a random order, playing at the speed of 0.01 second per frame. Participants were asked to watch entire video in a single take and provide time slots to generate video summary. We muted audio to ensure scores are based solely on visual stimuli. We have 25 different annotations for each video, taking more than 150 hours. Following \cite{Gygli,Song}, we calculated average pair-wise F1-score among collected human ground truth. The F1-score is $0.31$, which is approximately close to that of SumMe and TVSum dataset ($0.31$ and $0.36$). Meanwhile, we computed Cronbach $\alpha$, which is a standard measure to assess the reliability of a psychometric test. The dataset has a mean of $\alpha = 0.78$. The minimum value is $0.64$, and the maximum value is $0.92$. The Cronbach $\alpha$ of SumMe and TVSum datasets are $0.71$ and $0.81$ respectively. Ideally $\alpha$ is around 0.9, while $\alpha > 0.7$ is considered acceptable in exploratory researches. Thus, with UGSum52 dataset, video summarization experiments can be carried out with confidence. We provide a full list of the 52 videos, including name, camera type, frame number, video length, average summary length, average F-score and Cronbach $\alpha$ in Supplementary Material.

\section{Proposed Framework}
\label{Proposed Framework}

The proposed framework consists of three main steps as shown in Figure \ref{Overview}, which inputs a user-generated video and output a video summary. We construct a graph where vertex corresponding to a frame and edge between two vertexes is the KL divergence of two frames semantic probability distributions. First, we use a bundling center clustering algorithm to group video frames into disjoint segments of semantically consistent frames \cite{LeiZhuo}. Second, we rank segments according to the amount of semantic information contained in their frames using the graph-based FrameRank. Finally, we apply a greedy selection strategy to generate final summaries.

\subsection{Video Temporal Segmentation}
\label{segmentation}

Comparing to \cite{LeiZhuo}, we divide videos with a temporal segmentation method based on the clustering of deep semantic affinity graph. In general, connection between frames can be considered as a graph, where vertexes refers to the video frames and edges are the pairwise similarities.

The initial idea was to transfer a video to text and perform text summarization. However, image tagging and image caption are still open areas. Even with correctly classified labels, it is still difficult to precisely represent semantic content of video frames. Besides, frame contents may be different even though they have the same labels. To convey semantic information effectively, we decide to use the probability distribution of a set of labels to denote a frame. In detail, we feed video frames to a deep CNN \cite{Simonyan} pre-trained on images of 1000 object categories from ImageNet dataset \cite{Russakovsky} to compute the probabilities of frames containing objects. This representation enjoys the advantage of capturing information of the presence of a variety of object categories. 

We choose the KL divergence to measure how well a frame can represent another, which is a measurement of the difference between two probability distributions \cite{kldivergence}. It is the amount of information loss when a distribution is used to approximate another distribution, which can be interpreted how much a frame contains semantic information of another.

\begin{figure}[h]
\centering
\includegraphics[width=8cm]{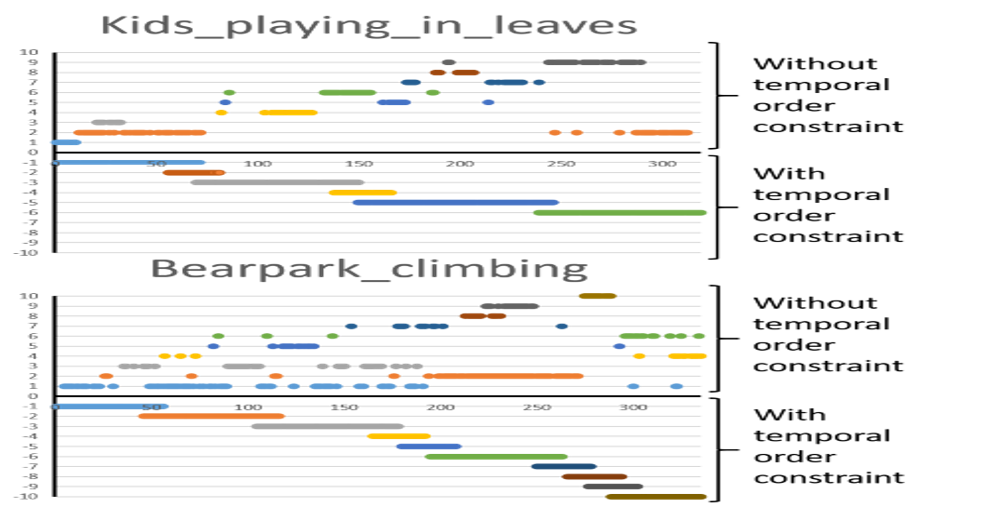}
\caption{Comparison between segments with and without using temporal constraint factor. Vertical axis is segment indexes, where the negative represent generated segment indexes with temporal order factor and the positive represent the ones without temporal order factor. Horizontal axis is temporal order.}
\label{time_compare}
\end{figure} 

We construct a graph $G(V,W)$, where $V=\{F_i\}$ are vertexes and $W=\{w_{ij}\}$ are edges between vertex $F_i$ and $F_j$. Edge $w_{ij}$ is the KL divergence computed as follows:

\begin{equation}
\label{kl divergence}
w_{ij}= - \sum_{k} P_{F_i(k)}\cdot \log \frac{P_{F_i(k)}}{P_{F_j(k)}}
\end{equation}
where $i$ and $j$ are frame indexes, and $\cdot$ means element-wise multiplication. $P_{F_i(k)}$ represents the probability of label $k$ of frame $F_i$. We negate values to transfer the difference into similarity and normalize matrix $G$. In addition, $G^\theta(V,W^\theta) $ is a constrained graph with a Gaussian function to maintain temporal order and smooth frame difference, where $W^\theta=\{w^\theta_{ij}\}$ are edges between frames. Each vertex can be represented as:

\begin{equation}
\label{GSF}
w^\theta_{ij} = \frac{1}{\sigma \sqrt{2\pi }} e^{-\frac{(i-j)^{2}}{2\sigma^{2}}}
\end{equation}
where $\sigma$ is a control parameter to modify temporal penalization and smoothness level. Hence a temporally constrained graph $G^\theta_{tc}$ can be represented as:

\begin{equation}
\label{toget}
G^\theta_{tc} = G \cdot G^\theta
\end{equation}

Furthermore, cluster center can be multiple similar frames rather than a single one, which is denoted as bundling center. With a dense-neighbor-based clustering method \cite{Qian2}, we can identify local clusters based on the edge connectivity on $G^\theta_{tc}$. To be noted, elements of a local cluster are locally similar to all of the other elements inside neighborhood instead of being close to a single element. More details can be referred to \cite{Qian2}. We show examples of the comparison between results with or without temporal constraint factor in Figure \ref{time_compare}.

\subsection{Segment Selection with FrameRank}
\label{selection}

The difficulty in video summarization is how to define important frames or segments to compose summary. There are no standard criteria for measuring the importance of video segments, even human subjects cannot agree on a universal basis. A good summary should be concise and retain the most informative and significant contents. In other words, selected frames or segments that compose the summary should be able to represent the unselected ones as much as possible. In this paper, we define important frames as the ones can substitute others with least information loss.

As described in Section \ref{segmentation}, with $G(V,W)$, we develop the FrameRank method which works similarly to the TextRank text ranking method \cite{TextRank} in natural language processing. We build a graph with its vertices corresponding to video frames and edges measure the similarity between the frames. We then implement a graph ranking technique to measure relative importance of each video frame as well as segment. We calculate importance score of the vertex $F_i$ as:

\begin{equation}
\label{framerank}
I(F_i)=(1-d)+d*\sum_{F_j\in In(F_i)}\frac{w_{ji}}{\sum_{F_l\in Out(F_j)}w_{jl}}I(F_j)
\end{equation}
where $d$ is a damping factor and $0 \leq d \leq 1$, which plays the role of integrating the model into the probability of jumping from a given vertex to another random vertex in the graph \cite{TextRank}. In detail, damping factor $d$ can be interpreted as a random suffer changing of visual contents which may be caused by a sudden camera moving in the case of user-generated videos. Following \cite{TextRank}, we set $d=0.85$. 

The running of the algorithm starts with arbitrary values assigned to each vertex in the graph, and iterates until convergence. In our implementation, we stop the iteration when importance scores between two consecutive iterations is below a given threshold. Let $I^k(F_i)$ be the score of vertex $F_i$ at iteration $k$, the iteration stops at $kth$ iteration if $\|(I^{(k)}(F_i)-I^{(k-1)}(F_i)\| \leq \epsilon $, where $\epsilon$ is a pre-set threshold. After the algorithm converges, each vertex has a score representing the importance of video frames associated with the vertex. The final importance scores of the vertices of FrameRank are not dependent on initial values, only the number of iterations to converge may be different.

Other related work \cite{Gygli} tried to estimate the score of segment by summing up all frame importance scores. However, it may result in longer segments getting larger importance scores. Thus, we compute relative importance score of the segment $S_n$ with the average importance:

\begin{equation}
\label{segmentscore}
I(S_n)=\frac{\sum_{i=t_{start}}^{t_{end}}I_{(F_i)}}{t_{start}-t_{end}+1}
\end{equation}
where $t_{start}$ and $t_{end}$ are start and end frames of the segment.

\subsection{Video Summary Generation}
\label{summarization}

We generate a video summary by selecting video segments that can substitute the others with the least information loss. Given the set of importance scores, we want to find a subset of segments with their total length below a pre-defined maximum $L$, while the total importance scores is maximized. In other words, we want to solve the optimization problem:

\begin{equation}
\label{summary}
\begin{array}{lcl} & max & \sum_{}^{}x_n I(S_n) \\ & s.t. & \sum_{}^{}x_n|S_n|\leq L\end{array}
\end{equation}
where $x_i \in \{0,1\}$ and $x_n = 1$ indicates the segment is selected. Under the assumption of independence between scores $I(S_n)$, this maximization is a standard 0/1-knapsack problem with a greedy selection strategy. Furthermore, user-generated videos rarely contain redundant interesting events, hence we do not account for redundancy and diversity.

\section{Evaluation and Discussion}
\label{results}

To demonstrate the effectiveness of our proposed video summarization approach, we evaluated and compared it with state-of-the-art methods. We carried on experiments on three video datasets, SumMe \cite{Gygli} ((1) \textbf{Interestingness} \cite{Gygli} (2) \textbf{Submodular} \cite{GygliCVPR15} (3) \textbf{dpp} \cite{ZhangKe} (4) \textbf{dppLSTM}\cite{KeZhangECCV2016} (5) \textbf{Video MMR} \cite{YingboLi}), TVSum \cite{Song} ((1) \textbf{Web Image Prior} \cite{Khosla} (2) \textbf{LiveLight} \cite{Khosla} (3) \textbf{dppLSTM}) and UGSum52 ((1) \textbf{Random} (2) \textbf{Uniform}). Following \cite{Gygli, Song}, we compute F-socre against human summaries for evaluation, according to temporal overlap comparing to computed ones.

\begin{table}[htbp]
\scriptsize
\centering
\caption{We compare our approach with state-of-the-art methods on SumMe, TVSum and UGSum datasets.}
\begin{tabular}{|p{2cm}|p{3cm}|p{2cm}|}
\toprule
\textbf{Dataset} & \textbf{Method} & \textbf{F-score} \\
\midrule
\multirow{7}[14]{*}{SumMe} & Interestingness & 0.394 \\
& Submodular & 0.397 \\
& dpp & 0.413 \\
& vsLSTM & 0.416 \\
& dppLSTM & 0.429 \\
& Video MMR & 0.266 \\
& FrameRank (ours) & \textbf{0.453 } \\
\midrule
\midrule
\multirow{6}[12]{*}{TVSum} & LiveLight & 0.460 \\
& Web Image Prior & 0.360 \\
& TVSum & 0.500 \\
& vsLSTM & 0.579 \\
& dppLSTM & 0.596 \\
& FrameRank (ours) & \textbf{0.601 } \\
\midrule
\midrule
\multirow{3}[6]{*}{UGSum} & Random & 0.161 \\
& Uniform + FrameRank & 0.322 \\
& FrameRank (ours) & \textbf{0.388 } \\
\bottomrule
\end{tabular}%
\label{max_result}%
\end{table}%

\subsection{Results}

Table \ref{max_result} summarizes the performance of our methods and contrasts to those attained by prior work. The highlighted numbers indicate FrameRank obtains the best performance in the corresponding setting. We achieve the highest overall F-score of $0.453$ on SumMe dataset and $0.601$ on TVSum dataset (the previous state-of-the-art published was $0.429$ and $0.596$ respectively \cite{KeZhangECCV2016}). Furthermore, we also carry on experiments on our UGSum52 dataset. Ours is superior to the other two methods where F-score is $0.388$. It shows not only the effectiveness of the proposed FrameRank method, but also the reliability of the proposed temporal segmentation method (by comparing FrameRank and Uniform + FrameRank methods). It proves our method is able to find important segments to produce an informative summary. The results demonstrate the proposed method can create video summaries closer to human-level performance than other methods. Meanwhile, it is interesting to see our result is better than all the supervised methods. We analyze there are no standard rules to define what important content is to summarize video. Thus, human generated summaries may be quite different from each other due to different human perception and personal experience. We believe training data for user-generated video summarization is not sufficient for supervised methods, and the generated model is not able to characterize the property to summarize videos. Moreover, since TVSum is slightly different from SumMe and UGSum52, where it only contains 10 categories of videos, Thus, in theory, the characteristic of TVSum should be suitable for supervised methods to learn video structure, while our FrameRank still performs better. Therefore, we have reasons to believe a good unsupervised method is more appropriate for user-generated video summarization.

\subsection{Analysis and Discussion}

\begin{figure}
\centering
\includegraphics[width=8cm]{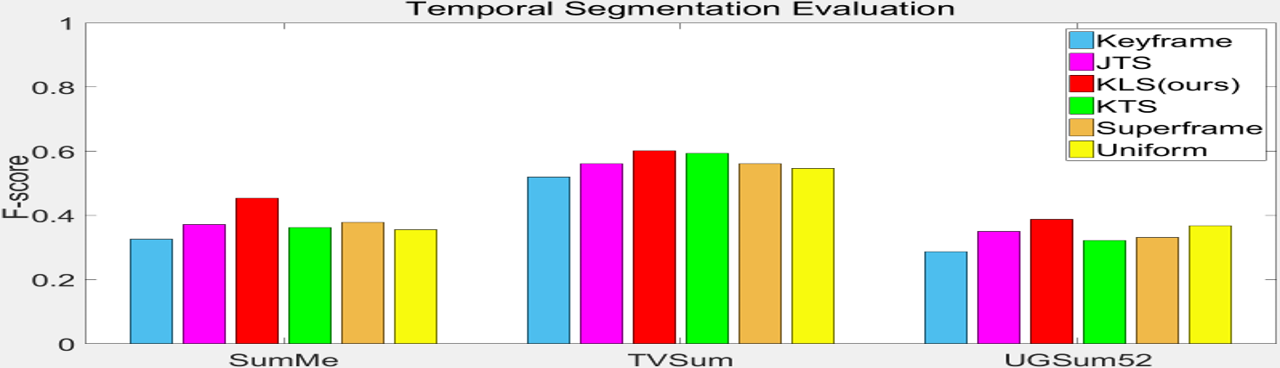}
\caption{Quantitative results with different features and similarity measurement.}
\label{fm}
\end{figure}

\begin{figure}
\centering
\includegraphics[width=8cm]{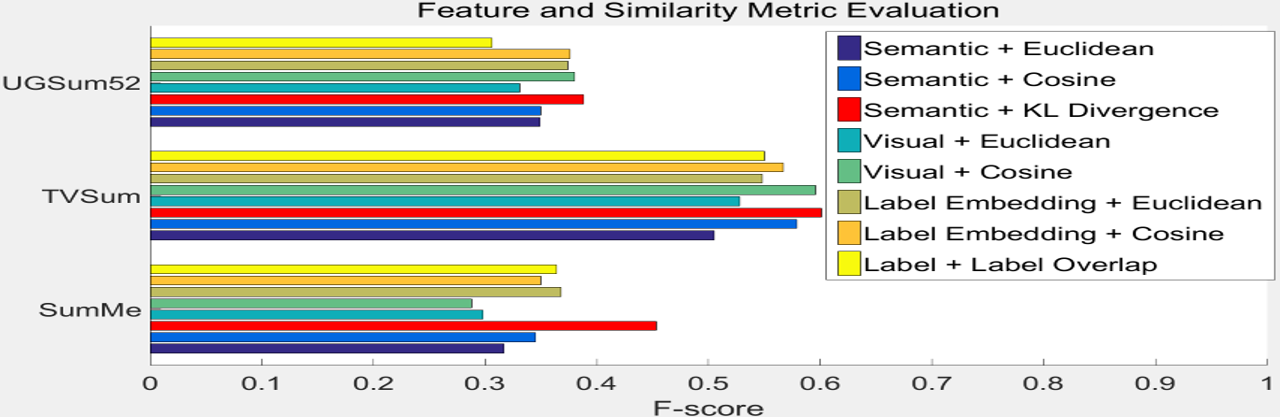}
\caption{Quantitative results of different temporal segmentation methods. KTS and JTS are short for kernel and joint-based temporal segmentation methods respectively.}
\label{combine}
\end{figure}

\textbf{Temporal video segmentation.} We analyze the performance gained by different temporal segmentation methods (Figure \ref{combine}). We compare our KL divergence based temporal segmentation approach with the following methods: 1) Joint temporal segmentation \textbf{(JTS)} \cite{LeiZhuo} 2) \textbf{Uniform} segmentation 3) Kernel-based temporal segmentation (\textbf{KTS}) \cite{Potapov} 4) \textbf{SuperFrame}, motion-based temporal segmentation \cite{Gygli} 5) \textbf{Keyframe} summarization, highest scored frames disregarding the temporal segment process. We employ the same segment selection method (FrameRank) and summary generation method.

Figure \ref{combine} shows our temporal segmentation method yields a better performance. It demonstrates the significance of structural analysis in video summarization. First, segment-based summarization is better than keyframes. We believe summarization annotation is segment-based rather than frame-based, because it is quite expensive and difficult for participants to give scores to individual frames. In fact, participants were not required to select a segment during annotation. Moreover, we find our approach has a greater advantage on SumMe and UGSum52 datasets, and close results for TVSum may be caused by similar videos. Thus, it demonstrates segment-based summary is in better agreement with human perception and produces more reasonable summaries because segments contain motion information comparing to keyframes. Furthermore, it also demonstrates our approach to cluster semantically similar frames matches better with human perception. It shows using such a grouping is indeed more semantically logical. Therefore, the experimental results prove the superiority of our segmentation approach, which is capable of generating meaningful summaries.

\textbf{Feature and similarity metrics.} We investigate the importance and reliability of different features and similarity metrics in the FrameRank approach. In Figure \ref{fm} we show the performance gained by using different features
(deep visual feature, deep semantic feature, label embedding and label) and different similarity metrics (Euclidean, Cosine, KL Divergence and Label Overlap). Label overlap is the TextRank. 

As could be expected, deep semantic features with the KL divergence metric performs the best. First, in most cases, features with semantic information are better than deep visual feature alone in summarizing video. Due to similar contents in TVSum, deep visual feature achieves relatively good results. Furthermore, it is interesting to observe label embedding feature has a comparable result to others (except semantic+KL).
Furthermore, the KL divergence has a better capability in measuring information loss when using one frame to represent others. Hence, it also proves our definition of importance of frames is reasonable and effective.

\section{Conclusion}
\label{conclusion}

We introduce a new benchmark - UGSum52 for user-generated video summarization. We have proposed a new unsupervised method for video summarization. With a novel dense-neighbor-based clustering method, our approach first partitions video into segments based on the deep semantic similarity of frames. We then develop a graph-based ranking method - FrameRank - to rank these segments. Finally, we sample segments with high information scores to generate video summary. We show our FrameRank method achieved results which are superiority to state-of-the-art methods.

\section{Acknowledgement}
\label{acknowledgement}

The author acknowledges the financial support from the International Doctoral Innovation Centre, Ningbo Education Bureau, Ningbo Science and Technology Bureau, and the University of Nottingham. This work was also supported by the UK Engineering and Physical Sciences Research Council [grant number EP/L015463/1].

\bibliographystyle{IEEEbib}
\bibliography{icme2019template}

\end{document}